\documentclass[conference]{IEEEtran}
\IEEEoverridecommandlockouts

\usepackage{cite}
\usepackage{amsmath,amssymb,amsfonts}
\usepackage[dvipdfmx]{graphicx}
\usepackage{textcomp}
\usepackage{xcolor}
\usepackage{multirow}

\usepackage[utf8]{inputenc} 
\usepackage[T1]{fontenc}    
\usepackage{hyperref}       
\usepackage{url}            
\usepackage{booktabs}       
\usepackage{amsfonts}       
\usepackage{nicefrac}       
\usepackage{microtype}      
\usepackage{longtable}

\usepackage{amsmath,amssymb,amsthm}

\DeclareMathOperator*{\minimize}{minimize}

\usepackage{algorithm}
\usepackage{algpseudocode}

\usepackage{graphicx}
\usepackage[caption=false]{subfig}
\usepackage{tabularx}

\usepackage{color}
\definecolor{dkgreen}{rgb}{0,0.6,0}
\definecolor{customgray}{rgb}{0.25,0.25,0.25}
\definecolor{customred}{rgb}{0.8,0.05,0.05}
\definecolor{customblue}{rgb}{0.05,0.05,0.8}

\usepackage{color}

\def\x{\mathbf{x}}
\def\m{\mathbf{m}}

\def\BibTeX{{\rm B\kern-.05em{\sc i\kern-.025em b}\kern-.08em
    T\kern-.1667em\lower.7ex\hbox{E}\kern-.125emX}}
\begin{document}

\title{Natural Evolution Strategy for Unconstrained\\ and Implicitly Constrained Problems\\ with Ridge Structure}

\author{\IEEEauthorblockN{Masahiro Nomura}
\IEEEauthorblockA{\textit{Tokyo Institute of Technology} \\
Yokohama, Japan \\
nomura.m.ad@m.titech.ac.jp}
\and
\IEEEauthorblockN{Isao Ono}
\IEEEauthorblockA{\textit{Tokyo Institute of Technology} \\
Yokohama, Japan \\
isao@c.titech.ac.jp}
}

\maketitle

\IEEEpubidadjcol

\begin{abstract}
In this paper, we propose a new natural evolution strategy for unconstrained black-box function optimization (BBFO) problems and implicitly constrained BBFO problems. 
BBFO problems are known to be difficult because explicit representations of objective functions are not available. 
Implicit constraints make the problems more difficult because whether or not a solution is feasible is revealed when the solution is evaluated with the objective function.
DX-NES-IC is one of the promising methods for implicitly constrained BBFO problems. DX-NES-IC has shown better performance than conventional methods on implicitly constrained benchmark problems. However, DX-NES-IC has a problem in that the moving speed of the probability distribution is slow on ridge structure. To address the problem, we propose the Fast Moving Natural Evolution Strategy (FM-NES) that accelerates the movement of the probability distribution on ridge structure by introducing the rank-one update into DX-NES-IC. The rank-one update is utilized in CMA-ES. Since naively introducing the rank-one update makes the search performance deteriorate on implicitly constrained problems, we propose a condition of performing the rank-one update. We also propose to reset the shape of the probability distribution when an infeasible solution is sampled at the first time. In numerical experiments using unconstrained and implicitly constrained benchmark problems, FM-NES showed better performance than DX-NES-IC on problems with ridge structure and almost the same performance as DX-NES-IC on the others. Furthermore, FM-NES outperformed xNES, CMA-ES, xNES with the resampling technique, and CMA-ES with the resampling technique.
\end{abstract}


\section{Introduction}

\IEEEPARstart{M}{any} real-world optimization problems are black-box and often have implicit constraints.
In black-box function optimization (BBFO), explicit representations of objective functions are not available, and only function evaluation values of solutions can be used.
Reducing the total computational budget is essential, as it is assumed that evaluating a solution requires large computational resources in many real-world BBFO problems. Hence, the goal of BBFO is to achieve better performance with as few function evaluations as possible.
Real-world optimization problems often have constraints that the solution must satisfy.
In implicitly constrained BBFO problems, which will be deliberately described in the next section, constraints are not given explicitly, and whether or not a solution is feasible is revealed when the solution is evaluated with the objective function.
In this paper, we consider unconstrained and implicitly constrained BBFO problems.

Natural Evolution Strategies (NES) \cite{glasmachers2010exponential,glasmachers2010natural,wierstra2014natural,beyer2014convergence,schaul2012natural,ollivier2017information,bensadon2015black} is a promising family for BBFO.
NES aims to minimize the expectation $J(\theta) = \int f({\bf x})p({\bf x}|\theta) d{\bf x}$ by updating the parameter $\theta^{*}$ of a probability distribution $p({\bf x}|\theta)$.
The update of NES is performed using the natural gradient~\cite{amari1998why,amari2000methods}.
The natural gradient is analytically intractable, then it is estimated by the Monte Carlo method.

DX-NES-IC~\cite{nomura2021distance} is a promising variant of NES.
A multivariate normal distribution is used as the probability distribution $p({\bf x}|\theta)$ in DX-NES-IC.
The experimental evaluations in \cite{nomura2021distance} show that DX-NES-IC shows better performance than DX-NES~\cite{fukushima2011proposal} and xNES~\cite{glasmachers2010exponential} on implicitly constrained problems.
While DX-NES-IC is originally proposed for improving the performance of DX-NES on \emph{implicitly constrained} problems, it has also succeeded in improving performance in \emph{unconstrained} ones~\cite{nomura2021distance}.
However, DX-NES-IC has a drawback that it moves slowly on problems with ridge structure.
We believe this is due to the lack of a mechanism that works with a small sample size in DX-NES-IC.
Actually, we have observed that DX-NES-IC shows worse performance than CMA-ES on some unconstrained problems with ridge structure such as the Rosenbrock function and the Cigar function.
More detailed examples and the importance of the ridge structure are explained in~\cite{whitley2004ruffled}.

To address the problem, we propose the Fast Moving Natural Evolution Strategy (FM-NES), which improves the search performance on problems with ridge structure.
To accelerate the movement on ridge structure, we propose to use the rank-one update, which stretches the multivariate normal distribution in the direction of movement.
The rank-one update is commonly used in the literature of covariance matrix adaptation evolution strategy (CMA-ES)~\cite{hansen2001completely};
however, using the rank-one update \emph{naively} causes performance deterioration in optimization for implicitly constrained problems.
To achieve efficient optimization on implicitly constrained problems, FM-NES tries to perform the rank-one update \emph{only} when the probability distribution is on ridge structure.
To do so, we introduce a condition of whether or not the probability distribution is on ridge structure.
In addition, FM-NES resets the shape of the covariance matrix when an infeasible solution is sampled at the first time.
With the reset operation, we expect that FM-NES avoids the negative effect of an elongated distribution due to the rank-one update.
We confirm the effectiveness of these operations in numerical experiments using benchmark problems.

The rest of the paper is organized as follows.
In Section~\ref{sec:problem_setting}, we explain the problem setting of this study.
In Section~\ref{sec:dxnesic}, we introduce DX-NES-IC and point out its problem.
In Section~\ref{sec:proposed}, we propose FM-NES, which addresses the problem of DX-NES-IC.
In Section~\ref{sec:experiments}, to demonstrate the effectiveness of FM-NES, we conduct numerical experiments using benchmark problems.
In Section~\ref{sec:discussion} , we discuss FM-NES in more detail by conducting some experiments.
In Section~\ref{sec:conclusion}, we conclude this paper.

\section{Problem Setting}
\label{sec:problem_setting}
In this paper, we focus on unconstrained BBFO problems and implicitly constrained BBFO ones~\cite{uemura2013new,uemura2016aega,nomura2021distance}.

The unconstrained BBFO problem is given by
\begin{align}
    \label{eq:bbfo}
    \minimize_{{\bf x} \in \mathbb{R}^d} f({\bf x}),
\end{align}
where $f$ is an objective function and is not explicitly given, and ${\bf x}$ is a solution.

On the other hand, the implicitly constrained BBFO problem is given by
\begin{align}
  \minimize_{{\bf x} \in \mathbb{R}^d} f_\mathrm{ic}({\bf x}) =
  \begin{cases}
    f({\bf x}), & {\bf x} \in \mathcal{S}, \\
    +\infty, & {\bf x} \notin \mathcal{S}, \\
  \end{cases}
\end{align}
where $\mathcal{S} \subset \mathbb{R}^d$ is a feasible region and is not given explicitly~\cite{uemura2016aega,nomura2021distance}.
In this setting, only the feasibility of a solution is revealed when the solution is evaluated, which means that the value of the constraint function cannot be used quantitatively. 
This property makes optimization on implicitly constrained problems more challenging.
Implicitly constrained problems often appear in real-world problems such as the lens system design~\cite{ono1998global,ono2000optimal} and the gene network estimation~\cite{ono2004evolutionary}. 

\section{DX-NES-IC and Its Problem}
\label{sec:dxnesic}

\subsection{DX-NES-IC}
The Distance-weighted eXponential Natural Evolution Strategy (DX-NES) taking account of Implicit Constraint (DX-NES-IC)~\cite{nomura2021distance} is a refined version of DX-NES.
DX-NES-IC iteratively updates the parameters of the multivariate normal distribution.
While DX-NES-IC is, as the name suggests, a method proposed for improving the performance of DX-NES on implicitly constrained problems, it has also succeeded in improving performance on unconstrained ones~\cite{nomura2021distance}.
The overall procedure of DX-NES-IC is shown in Algorithm~\ref{alg:dxnesic}.
We describe the detail of each step below.

\begin{algorithm}
\caption{Procedure of DX-NES-IC.}
\label{alg:dxnesic}
\begin{algorithmic}[1]
\State Initialize parameters.
\State Generate solutions.
\State Sort by preference order relation taking account of infeasible solutions.
\State Update evolution path.
\State Determine the search phase.
\State Switch weights according to the search phase.
\State Switch the learning rates according to the search phase.
\State Estimate the natural gradient.
\State Update the parameters by natural gradient.
\State Emphasize expansion of the distribution.
\State Check if the stopping criterion is met.
\end{algorithmic}
\end{algorithm}

{\bf Step 1.} The parameters of the multivariate normal distribution $\mathcal{N}({\bf m}^{(0)}, \sigma^{(0)} {\bf B}^{(0)}(\sigma^{(0)} {\bf B}^{(0)})^{\top})$ is initialized, where ${\bf m}^{(0)}$, $\sigma^{(0)}$, and ${\bf B}^{(0)}$ are the mean vector, the step size, and the normalized transformation matrix which holds ${\rm det}({\bf B}) = 1$ , respectively. The generation counter and the evolution path are set to $g = 0$ and ${\bf p}_{\sigma}^{(0)} = {\bf 0}$, respectively.

{\bf Step 2.} $\lambda$ solutions are generated as follows: ${\bf x}_{2i-1} = {\bf m}^{(g)} + \sigma^{(g)}{\bf B}^{(g)} {\bf z}_{2i-1}, {\bf x}_{2i} = {\bf m}^{(g)} + \sigma^{(g)} {\bf B}^{(g)} {\bf z}_{2i}, {\bf z}_{2i-1} \sim \mathcal{N} ({\bf 0}, {\bf I}), {\bf z}_{2i} = - {\bf z}_{2i-1} (i=1, \cdots, \lambda / 2)$, where $\lambda$ is a positive even number.

{\bf Step 3.} The generated solutions are sorted by the preference order relation. The definition of the preference order relation is as follows.
\begin{align}
{\bf x}_i \succ {\bf x}_j \Leftrightarrow
\begin{cases}
{\bf x}_i \in \mathcal{S} \wedge {\bf x}_j \in \mathcal{S} \wedge f({\bf x}_i) < f({\bf x}_j), \\
{\bf x}_i \in \mathcal{S} \wedge {\bf x}_j \notin \mathcal{S}, \\
{\bf x}_i \notin \mathcal{S} \wedge {\bf x}_j \notin \mathcal{S} \wedge \|{\bf z}_i\| < \|{\bf z}_j\|,
\end{cases} \label{eq:impord}
\end{align}
where ${\bf x}_i \succ {\bf x}_j$ means that ${\bf x}_i$ is better than ${\bf x}_j$.

{\bf Step 4.} The evolution path~\cite{hansen1996adapting,hansen2006cma} ${\bf p}_{\sigma}^{(g)}$ is updated as follows.
\begin{align}
\label{p_sigma_update}
{\bf p}_{\sigma}^{(g+1)} &= (1 - c_{\sigma}) {\bf p}_{\sigma}^{(g)} + \sqrt{c_{\sigma}(2 - c_{\sigma})\mu_{\mathrm{eff}}} \sum_{i=1}^{\lambda} w_i^{\mathrm{rank}} {\bf z}_{i:\lambda},
\end{align}
where ${\bf z}_{i:\lambda}$ denotes that ${\bf x}_i$ is the $i$-th best solution of the $\lambda$ ones.
The learning rate $c_{\sigma}$, the weight $w_i^{\rm rank}$, and $\mu_{\rm eff}$ are given by
\begin{align}
c_\sigma &= \frac{\mu_{\mathrm{eff}} + 2}{d + \mu_{\mathrm{eff}} + 5}, \\
w_i^{\mathrm{rank}}  &= \frac{\hat{w}_i^{\mathrm{rank}}} {\sum_{j=1}^\lambda \hat{w}_j^{\mathrm{rank}}} - \frac{1}{\lambda}, \label{eq:weight-rank} \\
\hat{w}_i^{\mathrm{rank}} &= \max \left(0, \ln\left(\frac{\lambda}{2} + 1 \right) - \ln{(i)} \right), \\
\label{eq:mueff}
\mu_{\mathrm{eff}} &= 1 / \sum_{i=1}^{\lambda}\left( w_i^{\mathrm{rank}} + \frac{1}{\lambda}\right)^2.
\end{align}

{\bf Step 5.} If $\mathbb{E} [\| \mathcal{N} ({\bf 0}, {\bf I}) \|] \leq \|{\bf p}_\sigma^{(g+1)} \|$, the search phase is set to ``movement''. 
If $0.1 \mathbb{E} [\| \mathcal{N} ({\bf 0}, {\bf I}) \|] \leq \|{\bf p}_\sigma^{(g+1)} \| < \mathbb{E} [\| \mathcal{N} ({\bf 0}, {\bf I}) \|]$, the search phase is set to ``stagnation''.
Otherwise, the search phase is set to ``convergence''.

{\bf Step 6.} If the search phase is ``movement'', the weight is set to $w_i = w_i^{\rm dist}$. Otherwise, the weight is set to $w_i = w_i^{\rm rank}$.
The distance weight $w_i^{\rm dist}$ is defined by
\begin{align}
\label{eq:weight-dist}
w_i^{\mathrm{dist}}  &= \frac{\hat{w}_i^{\mathrm{rank}}\hat{w}_i^{\mathrm{dist}}}{\sum_{j=1}^\lambda \hat{w}_j^{\mathrm{rank}}\hat{w}_j^{\mathrm{dist}}} - \frac{1}{\lambda}, \\
\label{w_dist_hat}
\hat{w}_i^{\mathrm{dist}} &= \exp\left(\alpha \| {\bf z}_i \| \right),
\end{align}
\noindent
where $\alpha$ is the distance weight parameter. 

{\bf Step 7.} If the search phase is ``movement'', the learning rates are set to $\eta_{\sigma} = \eta_{\sigma}^{\mathrm{move}}$ and $\eta_{\bf B} = \eta_{\bf B}^{\mathrm{move}}$.
If it is ``stagnation'', $\eta_{\sigma} = \eta_{\sigma}^{\mathrm{stag}}$ and $\eta_{\bf B} = \eta_{\bf B}^{\mathrm{stag}}$.
If it is ``convergence'', $\eta_{\sigma} = \eta_{\sigma}^{\mathrm{conv}}$ and $\eta_{\bf B} = \eta_{\bf B}^{\mathrm{conv}}$.
$\eta_{\bf m}$ is always set to one.
The recommended values of these learning rates are in~\cite{nomura2021distance}.

{\bf Step 8.} The natural gradients ${\bf G}_{\bf M}, G_{\sigma}, {\bf G}_{\bf B}$, and ${\bf G}_{\delta}$ are estimated:
\begin{align}
  \begin{split}
  {\bf G}_{\bf M} &= \sum_{i=1}^{\lambda} w_{i} ({\bf z}_{i:\lambda} {\bf z}_{i:\lambda}^{\top} - {\bf I}), G_{\sigma} = \mathrm{tr} ({\bf G}_{\bf M}) / d, \\
  {\bf G}_{\bf B} &= {\bf G}_{\bf M} - G_{\sigma} {\bf I}, {\bf G}_{\delta} = \sum_{i=1}^{\lambda} w_{i}{\bf z}_{i:\lambda},
  \end{split}
\end{align}
where ${\bf I} \in \mathbb{R}^{d \times d}$ is an identity matrix.

{\bf Step 9.} The parameters of the multivariate normal distribution are updated:
\begin{align}
  \label{sigma_update}
  \sigma^{(g+1)} &= \sigma^{(g)} \exp (\eta_{\sigma} G_{\sigma}/2), \\
  \label{m_update}
  {\bf m}^{(g+1)} &= {\bf m}^{(g)} + \eta_{\bf m} \sigma^{(g)} {\bf B}^{(g)} {\bf G}_{\delta}, \\
  \label{B_update}
  {\bf B}^{(g+1)} &= {\bf B}^{(g)} \exp(\eta_{\bf B} {\bf G}_{\bf B}/2).
\end{align}

{\bf Step 10.} When the search phase is ``movement'', emphasizing the expansion of the distribution is performed:
\begin{align}
  \label{eq:expand}
  {\bf Q} &= (\gamma-1)\sum_{i=1}^d {\mathbb I}(\tau_i > 0) {\bf e}_i {\bf e}_i^{\top} + {\bf I}, \\
  \label{expand-B}
  {\bf B}^{(g+1)} &\gets \mathbf{Q} {\bf B}^{(g+1)}/ \sqrt[d]{{\rm det}({\bf Q})}, \\
  \label{expand-sigma}
  \sigma^{(g+1)} &\gets \sigma^{(g+1)}\sqrt[d]{{\rm det}({\bf Q})},
\end{align}
where $\mathbf{Q} \in \mathbb{R}^{d \times d}$ is a matrix which expands the normalized transformation one and $\gamma$ is an expansion rate.
And, $\tau_i (i=1, \ldots, d)$ is a second moment change in each direction from ${\bf B}^{(g)} {\bf B}^{(g)^{\top}}$ to ${\bf B}^{(g+1)} {\bf B}^{(g+1)^{\top}}$, which is calculated as follows.
\begin{align}
  \tau_i &= \frac{ {\bf e}_i^{\top}({\bf B}^{(g+1)} {\bf B}^{(g+1)^{\top}} - {\bf B}^{(g)} {\bf B}^{(g)^{\top}}) {\bf e}_i}{ {\bf e}_i^{\top} {\bf B}^{(g)} {\bf B}^{(g)^{\top}} {\bf e}_i} \\
         &= \frac{ {\bf e}_i^{\top} {\bf B}^{(g+1)} {\bf B}^{(g+1)^{\top}} {\bf e}_i}{ {\bf e}_i^{\top} {\bf B}^{(g)} {\bf B}^{(g)^{\top}} {\bf e}_i} - 1 \quad (i \in \{ 1, \cdots, d \} ),
\end{align}
where ${\bf e}_i, i \in \{ 1, \cdots, d \} $ are the eigenvectors of the normalized covariance matrix ${\bf B}^{(g)} {\bf B}^{(g)^{\top}}$.
This makes the probability distribution expand in a direction of ${\bf e}_i$ if $\tau_i>0$.
The expansion ratio $\gamma$ is updated as follows.
\begin{align}
  \tau &= \max_i \tau_i, \\
  \gamma &\gets \max \Bigl( (1 - c_\gamma)\gamma + c_\gamma \sqrt{1+d_\gamma\tau}, 1\Bigr),
\end{align}
where $d_\gamma \in \mathbb{R}$ and $c_\gamma \in [0, 1]$.

{\bf Step 11.} The generation counter is updated $g \gets g+1$ and go back to Step 2 if the stopping criterion is not met.

\subsection{Problem of DX-NES-IC}
The performance of DX-NES-IC deteriorates when applied to problems with ridge structure.
We believe that this is because stretching the distribution in the direction of movement is slow on ridge structure.

\section{Proposed Method}
\label{sec:proposed}
In this section, we propose the \emph{Fast Moving Natural Evolution Strategy} (FM-NES), which addresses the problem of DX-NES-IC.

\subsection{Motivation}
To improve the search efficiency of DX-NES-IC on problems with ridge structure, we try to stretch the covariance matrix of the multivariate normal distribution in the moving direction by using the \emph{rank-one update}.
The rank-one update is also used in CMA-ES~\cite{hansen2001completely}.

However, it is observed that naively introducing the rank-one update into DX-NES-IC causes performance deterioration on implicitly constrained problems.

To alleviate the negative effect of the rank-one update on implicitly constrained problems, FM-NES tries to perform the rank-one update \emph{only} when the probability distribution is on ridge structure.
To do so, we introduce a condition of whether or not the probability distribution is on ridge structure.
If the probability distribution is on ridge structure, we believe that the rank-one update is still helpful even on implicitly constrained problems.
In this case, FM-NES performs the rank-one update as usual.
On the other hand, if it cannot be determined that the probability distribution is on ridge structure, FM-NES does not perform the rank-one update.
In addition, FM-NES resets the shape of the covariance matrix when an infeasible solution is sampled at the first time to cancel the negative effect of the rank-one update.

\subsection{Rank-One Update}
\label{sec:proposed_rankone}

To learn the direction of movement of the distribution, we first update the evolution path ${\bf p}_c$:
\begin{align}
    {\bf p}_{c}^{(g+1)} = (1-c_{c}) {\bf p}_{c}^{(g)} + \sqrt{c_{c}(2-c_{c})\mu_{\mathrm{eff}}} {\bf B}^{(g)}{\bf G}_{\delta},
\end{align}
where $c_c = (4 + \mu_{\rm eff}/d) / (d + 4 + 2 \mu_{\rm eff}/d)$ is a user parameter whose value is recommended in \cite{hansen2016cma}.
We then calculate the rank-one update over \emph{normalized} space, i.e., 
\begin{align}
    {\bf R} = ({{\bf B}^{(g)}}^{-1} {\bf p}_{c}^{(g+1)}) ({{\bf B}^{(g)}}^{-1} {\bf p}_{c}^{(g+1)})^{\top} - {\bf I}.
\end{align}
We then extract only the changes in the shape of the covariance matrix as follows:
\begin{align}
    \label{eq:only_shape}
    {\bf R}_{\bf B} = {\bf R} - \frac{{\rm Tr}({\bf R})}{d} {\bf I}.
\end{align}

Finally, the rank-one update is performed via exponential parameterization~\cite{glasmachers2010exponential,krause2015cma}:
\begin{align}
    \label{eq:proposed_rankone}
    {\bf B}^{(g+1)} &= {\bf B}^{(g)} \exp(c_1 {\bf R}_{\bf B} / 2),
\end{align}
where $c_1 \in \mathbb{R}$ is a learning rate.

This is closely related to the rank-one update used in xCMA-ES~\cite{krause2015cma}, which changes the volume of the multivariate normal distribution.
In contrast, the proposed rank-one update does not, i.e., ${\rm det}({\bf B}^{(g+1)}) = {\rm det}({\bf B}^{(g)})$, which is achieved by Eq.~(\ref{eq:only_shape}), as shown below.

For a matrix ${\bf M}$ which satisfies ${\rm Tr}({\bf M}) = 0$, 
$\det({\bf B} \exp({\bf M})) = \det({\bf B}) \cdot \det(\exp({\bf M})) = \det({\bf B}) \cdot \exp({\rm Tr}({\bf M})) = \det({\bf B})$.
Note that ${\rm Tr} (c_1 {\bf R}_{\bf B} / 2) = 0$ because ${\rm Tr}({\rm Tr}({\bf R}) {\bf I} / d) = {\rm Tr}({\bf R})$.
Therefore, ${\rm det}({\bf B}^{(g+1)}) = {\rm det}({\bf B}^{(g)})$ holds.

\subsection{Condition of Using Rank-One Update}
We introduce a condition of whether or not the probability distribution is on ridge structure for implicitly constrained problems.
To do so, we use the ratio of the square root of the largest eigenvalue of the normalized covariance matrix ${\bf B} {\bf B}^{\top}$ and that of the second largest eigenvalue.
If the ratio is larger than a threshold $\beta (> 1)$, the distribution is judged to be on ridge structure.
Let $\lambda_1({\bf B} {\bf B}^{\top})$ and $\lambda_2({\bf B} {\bf B}^{\top})$ be the largest eigenvalue and the second largest one of ${\bf B} {\bf B}^{\top}$, respectively.
Then, the condition is defined as follows:
\begin{align}
    \sqrt{\frac{\lambda_1({\bf B}^{(g+1)} {{\bf B}^{(g+1)}}^{\top})}{\lambda_2({\bf B}^{(g+1)} {{\bf B}^{(g+1)}}^{\top})}} > \beta.
\end{align}

\subsection{Resetting Shape of Multivariate Normal Distribution}
When an infeasible solution is sampled at the first time, we reset the learning of the normalized transformation matrix ${\bf B}^{(g)}$ to make it an initial matrix ${\bf B}^{(0)}$, which is usually an identity matrix.
In conjunction with the reset of ${\bf B}^{(g)}$, the evolution paths ${\bf p}_{\sigma}^{(g)}, {\bf p}_{c}^{(g)}$, and the parameter $\gamma$ used in the operation of emphasizing expansion of the multivariate normal distribution are also initialized:
\begin{align}
\begin{aligned}
    \label{eq:reset}
    {\bf B}^{(g)} = {\bf B}^{(0)}, {\bf p}_{\sigma}^{(g)} = {\bf 0}, {\bf p}_{c}^{(g)} = {\bf 0}, \gamma = 1  \\
\end{aligned}
\end{align}

\begin{algorithm}
\caption{Procedure of FM-NES. ${\bf E}$ is the set of positive even numbers and $\Upsilon := \mathbb{E}[\|\mathcal{N}({\bf 0}, \mbox{\bf I})\|]$. For clarity, the changes compared to DX-NES-IC are marked in \textcolor{blue}{blue}.}
\label{alg:fmnes}
\begin{algorithmic}[1]
\Require $\lambda \in {\bf E}, {\bf m}^{(0)} \in \mathbb{R}^d, \sigma^{(0)} \in \mathbb{R}, {\bf B}^{(0)} \in \mathbb{R}^{d \times d}$
\State $g = 0$, $\gamma = 1$, ${\bf p}_{\sigma}^{(0)} = {\bf 0}$, \textcolor{blue}{${\bf p}_{c}^{(0)} = {\bf 0}$, $h_{\rm unconst} = {\rm True}$}.
\While{stopping criterion not met}
  \For{$i \in \{ 1, \cdots, \lambda / 2 \}$}
    \State ${\bf z}_{2i-1} \sim \mathcal{N}({\bf 0}, {\bf I}), {\bf z}_{2i} = - {\bf z}_{2i-1}$
    \State ${\bf x}_{2i-1} = {\bf m}^{(g)} + \sigma^{(g)} {\bf B}^{(g)} {\bf z}_{2i-1}, {\bf x}_{2i} = {\bf m}^{(g)} + \sigma^{(g)} {\bf B}^{(g)} {\bf z}_{2i} $
  \EndFor
  \State Evaluate the solutions and sort $\{ ( {\bf z}_i, {\bf x}_i)\}$ by Eq.~\eqref{eq:impord}
  
  \If{\textcolor{blue}{$h_{\rm unconst} \land \# \textit{infeasible solutions} \geq 1$}}
    \State \textcolor{blue}{${\bf B}^{(g)} = {\bf B}^{(0)}, {\bf p}_{\sigma}^{(g)} = {\bf 0}, {\bf p}_{c}^{(g)} = {\bf 0}, \gamma = 1$}
    \State \textcolor{blue}{$h_{\rm unconst} \gets \rm{False}$}
  \EndIf
  
  \State ${\bf p}_{\sigma}^{(g+1)} = (1-c_{\sigma}) {\bf p}_{\sigma}^{(g)} + \sqrt{c_{\sigma}(2-c_{\sigma})\mu_{\mathrm{eff}}} \sum_{i=1}^{\lambda}w_i^{\rm rank} {\bf z}_{i}$
  \State {\bf if} $ \|{\bf p}_{\sigma}^{(g+1)}\| \geq \Upsilon$ {\bf then} $w_i = w_i^{\rm dist}$ {\bf else} $w_i = w_i^{\rm rank}$
  \State {\bf if} $ \|{\bf p}_{\sigma}^{(g+1)}\| \geq \Upsilon$ {\bf then} $\eta_{\sigma} = \eta_{\sigma}^{\rm move}, \eta_{\bf B} = \eta_{\bf B}^{\rm move}$ {\bf else if} $ \|{\bf p}_{\sigma}^{(g+1)}\| \geq 0.1\Upsilon$ {\bf then}  $\eta_{\sigma} = \eta_{\sigma}^{\rm stag}, \eta_{\bf B} = \eta_{\bf B}^{\rm stag}$ {\bf else} $\eta_{\sigma} = \eta_{\sigma}^{\rm conv}, \eta_{\bf B} = \eta_{\bf B}^{\rm conv}$

  \State ${\bf G}_{\delta} = \sum_{i=1}^{\lambda}w_i {\bf z}_{i}, {\bf G}_{\bf M} = \sum_{i=1}^{\lambda} w_i({\bf z}_{i} {\bf z}_{i}^{\top}-{\bf I})$
  \State $G_{\sigma} = {\rm tr}({\bf G_{\bf M}})/d, {\bf G}_{\bf B} = {\bf G}_{\bf M} - G_{\sigma}{\bf I}$

  \State ${\bf m}^{(g+1)} = {\bf m}^{(g)} + \eta_{m}\sigma^{(g)} {\bf B}^{(g)}{\bf G}_{\delta}$
  \State $\sigma^{(g+1)} = \sigma^{(g)} \cdot \exp( \eta_{\sigma}/2 \cdot G_{\sigma}) $
  \State ${\bf B}^{(g+1)} = {\bf B}^{(g)} \exp(\eta_{\bf B}{\bf G}_{\bf B} / 2)$
  
  \State \textcolor{blue}{${\bf p}_{c}^{(g+1)} = (1-c_{c}) {\bf p}_{c}^{(g)} + \sqrt{c_{c}(2-c_{c})\mu_{\mathrm{eff}}} {\bf B}^{(g)}{\bf G}_{\delta}$}
  
  \State $\{ {\bf e}_1, {\bf e}_2, \cdots, {\bf e}_d\} = {\rm eig}({\bf B}^{(g)}{{\bf B}^{(g)}}^{\top})$
  \For{$i \in \{ 1, \cdots, d \}$}
    \State $\tau_i = \frac{{\bf e}_i^{\top}{\bf B}^{(g+1)} {{\bf B}^{(g+1)}}^{\top} {\bf e}_i}{ {\bf e}_i^{\top} {\bf B}^{(g)} {{\bf B}^{(g)}}^{\top} {\bf e}_i}-1$
  \EndFor
  \State $\tau = \max_i \tau_i$
  \State $\gamma \leftarrow \max ( (1-c_{\gamma})\gamma + c_{\gamma} \sqrt{1+d_{\gamma}\tau}, 1 )$
  \If{$ \|{\bf p}_{\sigma}^{(g+1)}\| \geq \Upsilon$}
  \State ${\bf Q} = (\gamma-1)\sum_{i=1}^d {\mathbb I} ( \tau_i > 0 ) {\bf e}_i {\bf e}_i^{\top} + {\bf I}$
  \State $\sigma^{(g+1)} \leftarrow \sqrt[d]{{\rm det}({\bf Q})} \sigma^{(g+1)}$
  \State ${\bf B}^{(g+1)} = {\bf Q} {\bf B}^{(g+1)} / \sqrt[d]{{\rm det}({\bf Q})}$
  \EndIf
  \If{\textcolor{blue}{$h_{\rm unconst} \lor \sqrt{\frac{\lambda_1({\bf B}^{(g+1)} {{\bf B}^{(g+1)}}^{\top})}{\lambda_2({\bf B}^{(g+1)} {{\bf B}^{(g+1)}}^{\top})}} > \beta$}}
  \State \textcolor{blue}{${\bf R} = ({{\bf B}^{(g)}}^{-1} {\bf p}_{c}^{(g+1)}) ({{\bf B}^{(g)}}^{-1} {\bf p}_{c}^{(g+1)})^{\top} - {\bf I}$}
  \State \textcolor{blue}{${\bf R}_{\bf B} = {\bf R} - \frac{{\rm Tr}({\bf R})}{d} {\bf I}$}
  \State \textcolor{blue}{${\bf B}^{(g+1)} = {\bf B}^{(g+1)} \exp(c_1 {\bf R}_{\bf B} / 2)$}
  \EndIf

  \State $g \leftarrow g + 1$
\EndWhile

\end{algorithmic}
\end{algorithm}

\subsection{Algorithm of FM-NES}
\label{sec:fmnes-alg}
The overall procedure of FM-NES is shown in Algorithm~\ref{alg:fmnes}.
In lines 3-6, solutions are sampled from the multivariate normal distribution.
In line 7, the solutions are evaluated and sorted by using Eq.\eqref{eq:impord}.
In line 8-11, resetting of the shape of the covariance matrix and related parameters is performed when an infeasible solution is sampled at the first time.
In line 12, the evolution path ${\bf p}_{\sigma}$ is updated.
In lines 13-14, based on the norm of the evolution path, the weights and the learning rates are determined.
In lines 15-19, by using the estimated natural gradient, the parameters of the multivariate normal distribution are updated.
In line 20, the evolution path ${\bf p}_c$ is updated.
In lines 21-31, emphasizing the expansion of the probability distribution is performed.
In lines 32-36, the rank-one update is performed if the proposed condition is satisfied.

In FM-NES, we set $c_1$ to the same value used in~\cite{hansen2016cma} and $\beta = 1.2$ based on our preliminary experiments.
For other user parameters, the same values used in DX-NES-IC are employed~\cite{nomura2021distance}.

\section{Experiments}
\label{sec:experiments}

To demonstrate the efficiency of FM-NES, we compare the performance between FM-NES and conventional methods, DX-NES-IC~\cite{nomura2021distance}, xNES~\cite{glasmachers2010exponential}, and CMA-ES~\cite{hansen2014principled,hansen2016cma}, using benchmark problems. We also use xNES with the resampling technique (xNES/R) and CMA-ES with the resampling technique (CMA-ES/R) for implicitly constrained problems. 
The resampling technique can be used in implicitly constrained problems to handle the constraint.
The solution generation is repeated until $\lambda$ feasible solutions are obtained.
We employed the implementation \cite{resampling} for CMA-ES and CMA-ES/R.

\subsection{Benchmark problems}
We use the 40-dimensional benchmark problems shown in Table \ref{tab:benchmark}, four unconstrained problems and four implicitly constrained problems.
The Sphere problem, the Ellipsoid problem, the Rosenbrock problem, and the Cigar problem are unconstrained ones.
The IC-Sphere problem, the IC-Ellipsoid problem, the IC-Rosenbrock problem, and the IC-Cigar problem are implicitly constrained ones.
The Rosenbrock problem, the Cigar problem, the IC-Rosenbrock problem, and the IC-Cigar problem have ridge structure that this study focuses on.

\begin{table*}
  \centering
  \caption{Benchmark problem definitions.
  The column of ``${\bf m}^{(0)}$, $\sigma^{(0)}$'' shows an initial mean vector and an initial standard deviation of a multivariate normal distribution. 
  The bold number represents a $d$-dimensional vector whose each element is the number itself.}
  \label{tab:benchmark}
  \begin{tabular}{l|l|c|c|l}
    \hline
    Name & Objective function & Feasible region $\mathcal{S}$ & Optimum ${\bf x}^*$ & ${\bf m}^{(0)}$, $\sigma^{(0)}$ \\
    \hline \hline
    Sphere & $f({\bf x}) = \sum_{i=1}^{d}x_i^2$ & $\mathcal{S} = \mathbb{R}^d$ & $\mathbf{0}$ & ${\bf m}^{(0)}=\mathbf{20}, \sigma^{(0)}=2$ \\
    Ellipsoid & $f(\x) = \sum_{i=1}^{d}(1000^{\frac{i-1}{d-1}}x_i)^2$ & $\mathcal{S} = \mathbb{R}^d$ & $\mathbf{0}$ & $\m^{(0)}=\mathbf{20}, \sigma^{(0)}=2$ \\
    Rosenbrock & $f(\x) = \sum_{i=1}^{d-1}\left(100(x_{i+1} - x_i^2)^2 + (x_i - 1)^2\right)$ & $\mathcal{S} = \mathbb{R}^d$ & $\mathbf{1}$ & $\m^{(0)}=\mathbf{0}, \sigma^{(0)}=0.5$  \\    
    Cigar & $f(\x) = x_1^2 + \sum_{i=2}^{d}(100x_i)^2$ & $\mathcal{S} = \mathbb{R}^d$ & $\mathbf{0}$ & $\m^{(0)}=\mathbf{20}, \sigma^{(0)}=2$ \\
    IC-Sphere & $f({\bf x}) = \sum_{i=1}^{d}x_i^2$ & $x_i \geq 0, i=1, \ldots, d$ & $\mathbf{0}$ & ${\bf m}^{(0)}=\mathbf{20}, \sigma^{(0)}=2$ \\
    IC-Ellipsoid & $f(\x) = \sum_{i=1}^{d}(1000^{\frac{i-1}{d-1}}x_i)^2$ & $x_i \geq 0, i=1, \ldots, d$ & $\mathbf{0}$ & $\m^{(0)}=\mathbf{20}, \sigma^{(0)}=2$ \\
    IC-Rosenbrock & $f(\x) = \sum_{i=1}^{d-1}\left(100(x_{i+1} - x_i^2)^2 + (x_i - 1)^2\right)$ & $x_i \leq 1, i=1, \ldots, d$ & $\mathbf{1}$ & $\m^{(0)}=\mathbf{0}, \sigma^{(0)}=0.5$  \\
    IC-Cigar & $f(\x) = x_1^2 + \sum_{i=2}^{d}(100x_i)^2$ & $x_i \geq 0, i=1, \ldots, d$ & $\mathbf{0}$ & $\m^{(0)}=\mathbf{20}, \sigma^{(0)}=2$ \\
    \hline
  \end{tabular}
\end{table*}

\subsection{Performance metrics}
In the experiments, we regard the number of success out of 50 trials as the performance metrics.
Trials are terminated as successful when the evaluation value better than $10^{-10}$ is reached within the number of evaluations which does not exceed the pre-defined available number of evaluations. 
As the performance metrics, we also use the average of number of evaluations required for success if the number of success is the same.
The number of evaluations contains the number of evaluating infeasible solutions, as evaluating the solution is needed to confirm the feasibility in implicitly constrained problems.

\subsection{Experimental setups}
Based on results of preliminary experiments, we employed the one with the best performance metrics among 4, 8, 12, 16, 20, 24, 28, 32, 40, 60, and 80 as the sample size.
For CMA-ES and CMA-ES/R, the recommended value of the sample size $\lambda = 4+\lfloor 3\ln(d) \rfloor = 15$~\cite{hansen2006cma} was also investigated in the preliminary experiments.
Since we found that the sample size with the best performance of CMA-ES/R on the IC-Sphere problem and the IC-Cigar problem were 4 in the preliminary experiments, we also investigated the performance of the sample size of 2 and 3. As a result, we confirmed that CMA-ES/R showed better performance with the sample size of 4 than 2 and 3 on the IC-Sphere problem and the IC-Cigar problem. 
We set the available number of evaluations to $10^{6}$.

\subsection{Results}
The experimental results on the unconstrained and implicitly constrained problems are shown in Table \ref{tbl:ResultsOnUnconstrainedFunctions} and \ref{tbl:ResultsOnConstrainedFunctions}, respectively.
The results show that the performance of FM-NES is better than that of the other methods on the Ellipsoid problem, the Rosenbrock problem, the Cigar problem, the IC-Rosenbrock problem, and the IC-Cigar problem.
In particular, for the Rosenbrock problem, the Cigar problem, the IC-Rosenbrock problem, and the IC-Cigar problem, which have ridge structure, FM-NES has succeeded in finding the optimal solutions about 1.71, 1.78, 1.67, and 1.42 times faster than DX-NES-IC, respectively.
Furthermore, FM-NES showed better performance than CMA-ES on the unconstrained problems with ridge structure, the Rosenbrock problem and the Cigar problem, on which DX-NES-IC showed worse performance than CMA-ES.
On the other hand, on the Sphere problem, the IC-Sphere problem, and the IC-Ellipsoid problem, FM-NES shows almost the same performance as DX-NES-IC and better performance than xNES, CMA-ES, xNES/R, and CMA-ES/R.
Overall, the result suggests the effectiveness of FM-NES.

\begin{table*}[tb]
  \caption{Results of unconstrained problems.
  ``\#Suc.'' denotes the number of success.
  ``\#Eval.($\cdot$)'' denotes the average of number of evaluations and the standard deviation.
  }
  \label{tbl:ResultsOnUnconstrainedFunctions}
  \centering
  \begin{tabular}{c|c|c|c|c|c|c|c|c}
  \hline
  & \multicolumn{2}{c|}{Sphere} & \multicolumn{2}{c|}{Ellipsoid} & \multicolumn{2}{c|}{Rosenbrock} & \multicolumn{2}{c}{Cigar} \\ \hline
  Method & \#Suc. & \#Eval.($\times 10^{3}$) ($\lambda$) & \#Suc. & \#Eval.($\times 10^{3}$) ($\lambda$) & \#Suc. & \#Eval.($\times 10^{3}$) ($\lambda$) & \#Suc. & \#Eval.($\times 10^{3}$) ($\lambda$) \\
  \hline
  \hline
  FM-NES & 50 & $4.82\pm0.184$ (8) & 50 & $36.1\pm1.07$ (16) & 50 & $48.6\pm1.20$ (16) &  50 & $13.0\pm0.359$ (8) \\ \hline
  DX-NES-IC & 50 & $4.84\pm0.191$ (8) & 50 & $42.9\pm1.94$ (20) & 50 & $83.1\pm2.25$ (20) & 50 & $23.1\pm0.924$ (20) \\ \hline 
  xNES & 50 & $122\pm0.750$  (8) & 50 & $156\pm0.944$ (8) & 50 & $219\pm13.4$ (12) & 50 & $183\pm2.34$ (8) \\ \hline
  CMA-ES & 50 & $5.88\pm0.164$ (8) & 50 & $75.1\pm0.694$ (12) & 50 & $76.5\pm4.24$ (8) & 50 & $17.0\pm0.263$ (8) \\ \hline
  \end{tabular}
\end{table*}

\begin{table*}[tb]
  \caption{Results of implicitly constrained problems.
  ``\#Suc.'' denotes the number of success.
  ``\#Eval.($\cdot$)'' denotes the average of number of evaluations and the standard deviation.
  }
  \label{tbl:ResultsOnConstrainedFunctions}
  \centering
  \begin{tabular}{c|c|c|c|c|c|c|c|c}
  \hline
  & \multicolumn{2}{c|}{IC-Sphere} & \multicolumn{2}{c|}{IC-Ellipsoid} & \multicolumn{2}{c|}{IC-Rosenbrock} & \multicolumn{2}{c}{IC-Cigar} \\ \hline
  Method & \#Suc. & \#Eval.($\times 10^{3}$) ($\lambda$) & \#Suc. & \#Eval.($\times 10^{3}$) ($\lambda$) & \#Suc. & \#Eval.($\times 10^{3}$) ($\lambda$) & \#Suc. & \#Eval.($\times 10^{3}$) ($\lambda$) \\
  \hline
  \hline
  FM-NES & 50 & $19.3\pm1.17$ (12) & 50 & $159\pm9.10$ (60) & 50 & $69.9\pm1.48$ (20) &  50 & $63.0\pm3.26$ (20) \\ \hline
  DX-NES-IC & 50 & $19.6\pm1.38$ (12) & 50 & $164\pm10.2$ (60) & 50 & $117\pm3.50$ (24) & 50 & $89.5\pm3.75$ (20) \\ \hline 
  xNES & 50 & $636\pm2.45$  (40) & 50 & $798\pm2.93$ (40) & 50 & $389\pm3.62$ (28) & 47 & $927\pm2.82$ (40) \\ \hline
  xNES/R & 0 & $-$  ($-$) & 0 & $-$ ($-$) & 0 & $-$ ($-$) & 0 & $-$ ($-$) \\ \hline
  CMA-ES & 50 & $35.9\pm1.21$ (36) & 50 & $259\pm7.03$ (60) & 50 & $129\pm4.04$ (20) & 50 & $82.7\pm1.72$ (32) \\ \hline
  CMA-ES/R & 50 & $25.2\pm0.945$ (4) & 50 & $585\pm30.5$ (8) & 50 & $192\pm3.22$ (32) & 50 & $69.3\pm2.34$ (4) \\ \hline
  \end{tabular}
\end{table*}

\section{Discussions}
\label{sec:discussion}

\subsection{Effect of introducing rank-one update on problems with ridge structure}
In the previous section, FM-NES achieved faster convergence than DX-NES-IC on the problems with ridge structure.
In this section, we investigate the effect of the rank-one update by observing the more detailed behavior.
Specifically, for each method, we observe the transition of the square roots of the eigenvalues of the normalized covariance matrix ${\bf B} {\bf B}^{\top}$ and that of the best evaluation value, when each method is applied to a problem with ridge structure, the 40-dimensional Rosenbrock problem.
The initial distribution used in this experiment is the same as the one used in Section~\ref{sec:experiments}.
Based on the results of Table \ref{tbl:ResultsOnUnconstrainedFunctions}, the sample size of FM-NES and that of DX-NES-IC are set to 16 and 20, respectively.

\begin{figure*}[tb]
  \begin{center}
    \begin{tabular}{c}
      \begin{minipage}{0.48\hsize}
        \begin{center}
          \includegraphics[width=0.9\linewidth]{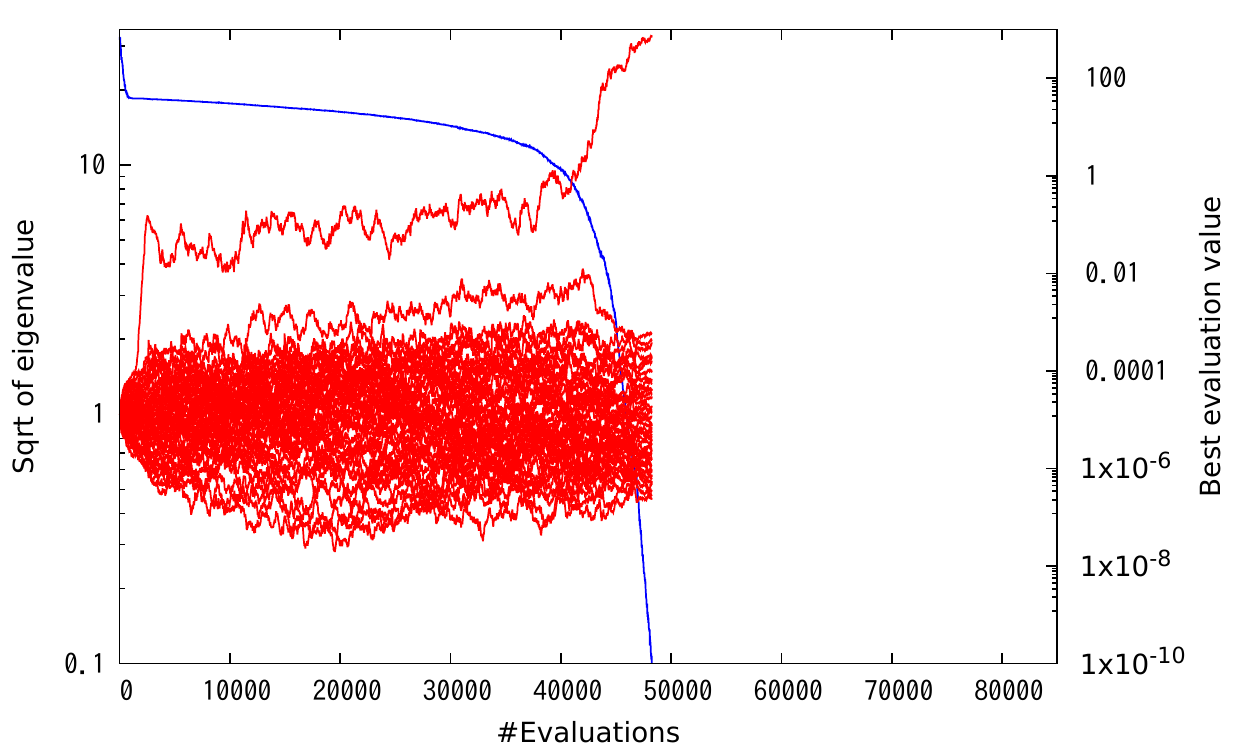}
          \hspace{2cm} {\small (a) FM-NES}
        \end{center}
      \end{minipage}
      \begin{minipage}{0.48\hsize}
        \begin{center}
          \includegraphics[width=0.9\linewidth]{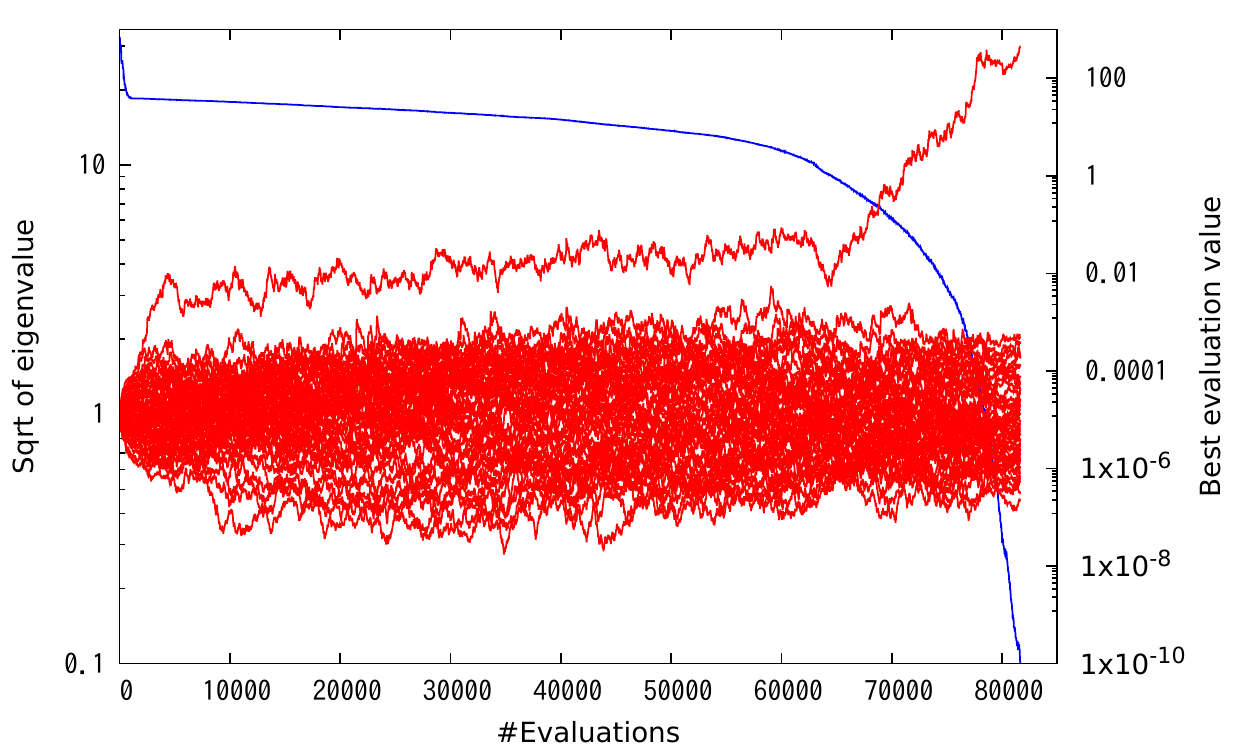}
          \hspace{2cm} {\small (b) DX-NES-IC}
        \end{center}
      \end{minipage}
    \end{tabular}
    \caption{The transition plots of the square roots of all eigenvalues of the normalized covariance matrix ${\bf B} {\bf B}^{\top}$ and the best evaluation value in a typical trial when FM-NES and DX-NES-IC are applied to the 40-dimensional Rosenbrock function. The horizontal axis, the left vertical one, and the right vertical one are the number of evaluations, the square root of eigenvalue, and the best evaluation value, respectively. The blue curves and the red ones are the best evaluation values and the square roots of the eigenvalues, respectively.}
    \label{fig:fval_and_eig}
  \end{center}
\end{figure*}

Figure \ref{fig:fval_and_eig} shows the transition of the square roots of the eigenvalues of ${\bf B} {\bf B}^{\top}$ and that of the best evaluation value in a typical trial of FM-NES and DX-NES-IC.
Comparing the both methods, the difference between the square root of the largest eigenvalue and that of the second largest one of FM-NES is larger than that of DX-NES-IC.
This result suggests that introducing the rank-one update makes the probability distribution stretch in the direction of movement of the probability distribution, which would contribute to the efficient movement of ridge structure. As a result, we believe that FM-NES was able to obtain the optimal solution faster than DX-NES-IC.

\subsection{Ablation study on the proposed rank-one update}
In implicitly constrained problems, FM-NES determines whether to perform the rank-one update by using the proposed condition of whether or not the probability distribution is on ridge structure.
To verify the effectiveness of the proposed condition, we compare the performance between FM-NES and a method named \emph{Method A}. Method A is the same as FM-NES except that it does not use the proposed condition and always performs the rank-one update.

FM-NES resets the learning of the normalized transformation matrix ${\bf B}$ to make it an initial matrix when an infeasible solution is sampled at the first time.
To investigate the effect of the reset, we compare the performance between FM-NES and a method named \emph{Method B}. Method B is the same as FM-NES except that it does not reset the learning of the normalized transformation matrix.

We also compare FM-NES and a method named \emph{Method C} which is the same as FM-NES except that it always performs the rank-one update and does not reset the learning of the normalized transformation matrix.

Table \ref{fig:DiscussionOnICSphere} shows the result on the 40-dimensional IC-Sphere problem. 
The experimental settings are the same as Section~\ref{sec:experiments}. 
From Table \ref{fig:DiscussionOnICSphere}, FM-NES shows the best performance of the four. 
The result suggests that the proposed condition and the reset of the learning of the normalized transformation matrix ${\bf B}$ are effective.

\begin{table}[tb]
\caption{Performance of FM-NES, Method A, Method B, and Method C on the IC-Sphere problem.}
\label{fig:DiscussionOnICSphere}  \centering
  \begin{tabular}{c|c|c}
  \hline
  Method & \#Suc. & \#Eval.($\times 10^{3}$) ($\lambda$) \\
  \hline
  \hline
  FM-NES & 50 & $19.3\pm1.17$ (12) \\ \hline
  Method A & 50 & $23.2\pm1.34$ (16) \\ \hline 
  Method B & 50 & $21.4\pm1.63$  (12) \\ \hline
  Method C & 50 & $24.3\pm1.43$  (16) \\ \hline
  \end{tabular}
\end{table}

\section{Conclusion}
\label{sec:conclusion}
In this work, we introduced a NES variant, the Fast Moving Natural Evolution Strategy (FM-NES) to improve the performance of DX-NES-IC on problems with ridge structure.
FM-NES accelerates the movement of the probability distribution on ridge structure by introducing the rank-one update into DX-NES-IC.
Since naively introducing the rank-one update makes the search performance deteriorate on implicitly constrained problems, we proposed the condition of performing the rank-one update.
We also proposed to reset the shape of the probability distribution when an infeasible solution is sampled at the first time.
In numerical experiments using unconstrained and implicitly constrained benchmark problems, FM-NES outperformed DX-NES-IC on problems with ridge structure and almost the same performance as DX-NES-IC on the others. Furthermore, FM-NES showed better performance than xNES, CMA-ES, and those with the resampling technique.

Future work would focus on applying FM-NES to real-world problems.
In particular, it is interesting to apply FM-NES to domains where CMA-ES has already been successfully applied~\cite{fujii2019topology,maki2020application,nomura2021warm}, and compare the performance of these methods.
Furthermore, extending FM-NES so that it can be applied to high-dimensional problems is an important direction to expand the scope of this method.

\bibliographystyle{IEEEtran}
\bibliography{ref}

\end{document}